\documentclass[10pt,twocolumn,letterpaper]{article}

\usepackage{cvpr}              
\usepackage{amssymb}
\usepackage{amsmath}
\usepackage{algorithmic}
\usepackage{algorithm}
\usepackage{graphicx}
\usepackage{wrapfig}
\usepackage{multirow}
\usepackage{multicol}
\usepackage{subcaption}
\usepackage{xfrac}
\usepackage{tabularx}

\newcommand{\R}[1]{{%
    \textbf{%
        \ifstrequal{#1}{1}{\textcolor{red}{R#1}}{%
        \ifstrequal{#1}{2}{\textcolor{blue}{R#1}}{%
        \ifstrequal{#1}{3}{\textcolor{magenta}{R#1}}{%
        \ifstrequal{#1}{4}{\textcolor{teal}{R#1}}{%
                           \textcolor{cyan}{R#1}%
        }}}}%
    }%
}}

\definecolor{cvprblue}{rgb}{0.21,0.49,0.74}
\usepackage[pagebackref,breaklinks,colorlinks,allcolors=cvprblue]{hyperref}

\def\paperID{11087} 
\def\confName{CVPR}
\def\confYear{2025}

\title{
Balanced Direction from Multifarious Choices: \\
Arithmetic Meta-Learning for Domain Generalization}

\author{
Xiran Wang$^{1}$ \qquad
    Jian Zhang$^{1}$ \qquad
    Lei Qi$^{2}$ \qquad
    Yinghuan Shi$^{1,3,}$\thanks{
     Xiran Wang, Jian Zhang, and Yinghuan Shi are with State Key Laboratory for Novel Software Technology, Nanjing University, China. This work was supported by National Science and Technology Major Project (2023ZD0120700), NSFC Project (62222604, 62206052, 624B2063), China Postdoctoral Science Foundation (2024M750424), Fundamental Research Funds for Central Universities (020214380120, 020214380128), State Key Laboratory Fund (ZZKT2024A14), Postdoctoral Fellowship Program of CPSF (GZC20240252), Jiangsu Funding Program for Excellent Postdoctoral Talent (2024ZB242), and Jiangsu Science and Technology Major Project (BG2024031). Corresponding: Yinghuan Shi.} \\
    $^{1}$Nanjing University \qquad
    $^{2}$Southeast University \qquad
    $^{3}$Suzhou Laboratory\\
    {\tt\small \{zzwdx, zhangjian7369\}@smail.nju.edu.cn, qilei@seu.edu.cn, syh@nju.edu.cn}
}

\begin{document}
\maketitle
\begin{abstract}

Domain generalization is proposed to address distribution shift, arising from statistical disparities between training source and unseen target domains. 
The widely used first-order meta-learning algorithms demonstrate strong performance for domain generalization by leveraging the gradient matching theory, which aims to establish balanced parameters across source domains to reduce overfitting to any particular domain.
However, our analysis reveals that there are actually numerous directions to achieve gradient matching, with current methods representing just one possible path.
These methods actually overlook another critical factor that the balanced parameters should be close to the centroid of optimal parameters of each source domain.
To address this, we propose a simple yet effective arithmetic meta-learning with arithmetic-weighted gradients.
This approach, while adhering to the principles of gradient matching, promotes a more precise balance by estimating the centroid between domain-specific optimal parameters. 
Experimental results validate the effectiveness of our strategy.
Our code is available at \href{https://github.com/zzwdx/ARITH}{https://github.com/zzwdx/ARITH}.

\end{abstract}

\section{Introduction}
\label{sec:intro}

\begin{figure}[t]
  \centering
  \includegraphics[width=1\linewidth]{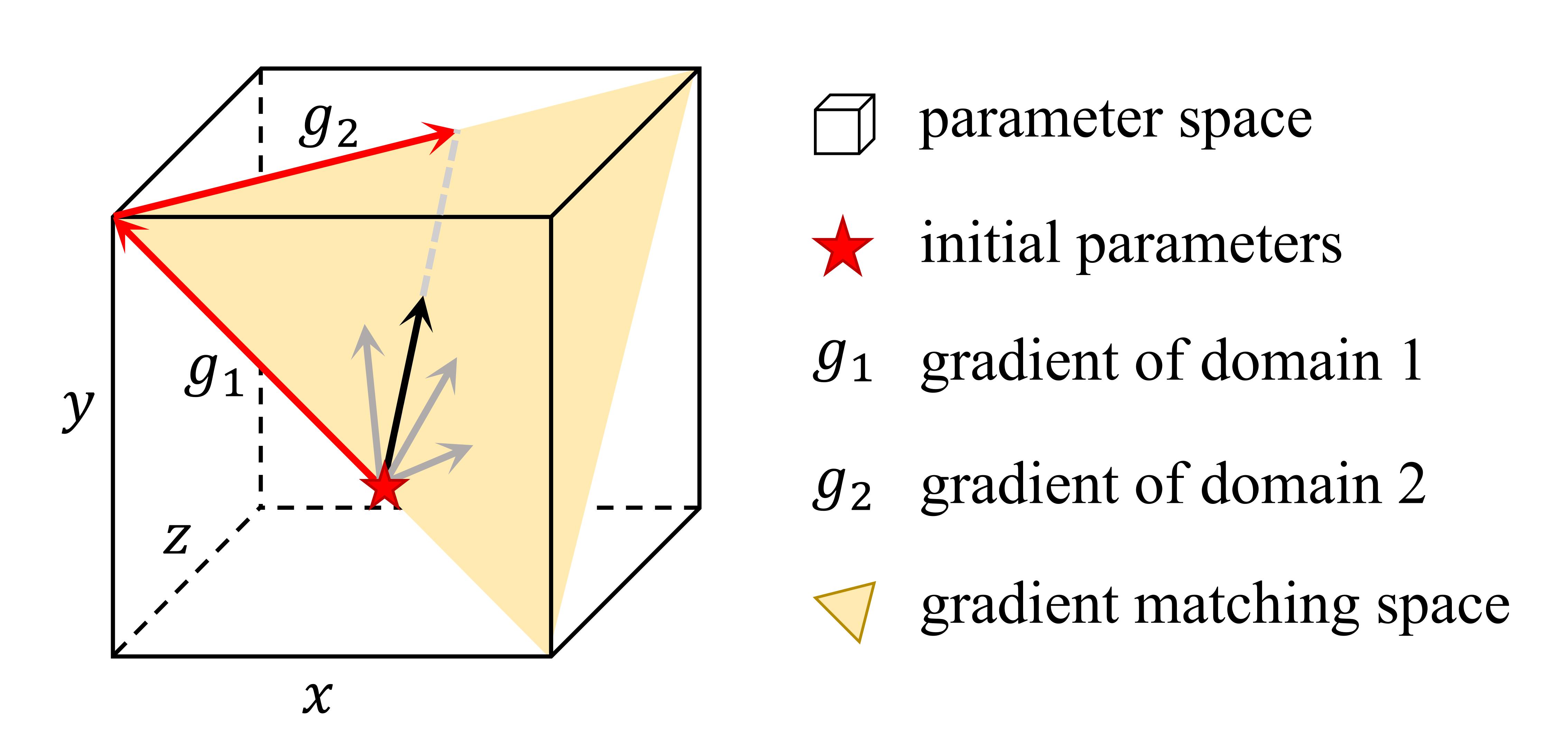}
   \caption{
   In the ternary parameter space $(x, y, z)$,
 gradient matching can be deduced in any directions of the yellow surface.
 The black arrow is the updating direction of existing methods, which moves away from the optimal solution of domain 1.
}
    \vspace{-0.1in}
    \label{fig:range}
\end{figure}
Deep neural networks \cite{krizhevsky2017imagenet} usually hinge on the premise that both training and test data are independent and identically distributed (i.i.d.).  However, this assumption often fails 
in dynamic real-world contexts, resulting in performance degradation
when test data diverges from the distribution encountered during training \cite{li2019research}. This shift in distribution proves detrimental, as the model tends to overfit specific representations that may be absent or inadequately represented during the test phase \cite{shah2020pitfalls, izmailov2022feature, qiu2024devil}.

In recent years, the domain generalization (DG) \cite{wang2022generalizing} paradigm has been widely studied to deal with distribution shift, 
which refers to leveraging multiple source domains to develop a model with the generalization ability that can be directly applied to arbitrary unseen target domains.

Some studies identify \emph{invariance} as a critical factor for domain generalization, which assume that certain stable elements persist across source domains and can also apply to unseen target domains.
These methods often employ techniques such as data augmentation \cite{zhou2020domain, li2021simple, guo2023aloft} and adversarial learning \cite{li2018domain, dayal2024madg} to capture invariant features, while others impose more intricate forms of invariance at gradient \cite{shahtalebi2021sand, wang2023sharpness} or predictor \cite{arjovsky2020invariant, krueger2021out} level.
However, the presumed invariance doesn't always translate to improved generalization due to the random nature of target domains. 
As illustrated in \cite{gulrajani2020search}, only a few of them surpass vanilla empirical risk minimization (ERM) \cite{naumovich1998statistical} in terms of average accuracy across multiple standard datasets.

Rather than strictly enforcing an invariance factor, \emph{balance}-based methods \cite{dou2019domain, cha2021swad, rame2022diverse} take a more flexible way by ensuring that model parameters are balanced across various domains. For example, meta-learning \cite{li2018learning, zhang2020adaptive, shu2021open} aims to achieve an optimal balance among source domains by implicitly guiding their gradient directions called \textbf{gradient matching}. The rationale is that large angles between gradients indicate conflicting objectives, suggesting that updating one domain may negatively impact the optimization process of others. In contrast, smaller gradient angles imply that optimizing one domain does not disrupt other domains, allowing for a mutually beneficial outcome by optimizing their combined gradient, thereby mitigating the risk of overfitting to specific domains \cite{shi2021gradient, wang2023generalizable}.

Although gradient matching currently serves as the basic theory in meta-learning for domain generalization, we believe it has not been fully explored.
As shown in \cref{fig:range}, each red arrow represents a gradient updating process. While existing methods often select the direction of $g_1 + g_2$, there are numerous other ways in the yellow region that can also achieve gradient matching.
Alternatively, the centroid of a structure is known for its balanced nature. 
In \cref{fig:surface} (d)(h), updating direction towards the centroid of source domain experts, marked by the yellow color, tend to lie closer to the target domain's loss basin \cite{izmailov2018averaging, cha2021swad} with better generalization ability than those in the $g_1 + g_2$ direction. 
This suggests that focusing only on gradient matching without considering \textbf{balanced positioning} seems insufficient.

We propose an arithmetic meta-learning framework by adjusting the weights of $g_1$ and $g_2$ in \cref{fig:range} to identify a more balanced position between source domains, where the weights of gradients are selected to form an arithmetic progression.
Intuitively, since $g_2$ is less correlated with the initial model than $g_1$, it is reasonable to assign $g_2$ with smaller weights.
We further demonstrate that a set of arithmetically decreasing gradient weights not only follows the principle of gradient matching, but also reflects model averaging \cite{izmailov2018averaging} that indirectly estimate the centroid of domain experts. This centroid approximation is expected to achieve a more accurate balance across source domains to enhance model's generalization stability in the unseen target scenarios.
Our contribution can be summarized as follows: 
\begin{itemize}
\item{We prove that existing first-order meta-learning strategies for domain generalization represent just one of many possible directions for gradient matching, and they overlook the need to balance the model's position from the optimal parameters of each source domain.}
\item{
We integrate model averaging into meta-learning and propose an arithmetic gradient-based strategy to simulate this process, which aims to estimate the centroid of domain-specific optimal parameters while ensuring consistency in gradient direction.
Our method is simple to implement, requiring adjustments of one line in \cref{alg:arithmetic}.
}
\item{
Arithmetic meta-learning outperforms traditional meta-learning strategies across multiple domain generalization benchmarks, and shows synergistic potential when integrated with global averaging techniques.
}

\end{itemize}

\begin{figure}[t]
\vspace{-0.05in}
\begin{algorithm}[H]
\caption{Arithmetic Meta-Learning for DG}  
\begin{algorithmic}[1]    
\label{alg:arithmetic}
\renewcommand{\algorithmicrequire}{\textbf{Input:}}
\REQUIRE Domains $\lbrace \mathcal{D}_1,\mathcal{D}_2,...,\mathcal{D}_S\rbrace$; model parametrized by $\Theta$; hyperparameters $k$ and $\epsilon$
\FOR{iterations $= 1,2,...$}
    \STATE Initialize parameters $\theta_1 \leftarrow \Theta$;
    \FOR{$\mathcal{D}_i \in \lbrace \mathcal{D}_1,\mathcal{D}_2,...,\mathcal{D}_S\rbrace$}
    \STATE Perform $k$ gradient updates from $\mathcal{D}_i$ to obtain $\theta_{i+1} \leftarrow \theta_i$ with $g_i = \theta_{i} - \theta_{i+1}$
    \ENDFOR
    \STATE \textcolor{gray}{$\Theta \leftarrow \Theta-\epsilon \sum_{i=1}^n g_i$ \, \texttt{// for Fish}}
    \STATE $\Theta \leftarrow \Theta-\frac{1}{n+\epsilon} \sum_{i=1}^n (n+1-i) g_i$ \, \texttt{\textcolor[rgb]{0.5, 0.5, 1}{// for Arith}}
    
\ENDFOR
\end{algorithmic}
\end{algorithm}
\vspace{-0.25in}
\end{figure}

\section{Method}
\label{sec:method}

\subsection{Preliminary}
\label{subsec:method-preliminary}

\textbf{Problem setting.} In domain generalization, we are provided with source domains $\mathcal{S} = \lbrace \mathcal{D}_1,\mathcal{D}_2,...,\mathcal{D}_S\rbrace$ and unseen target domains $\mathcal{T} = \lbrace \mathcal{D}_{S+1},\mathcal{D}_{S+2},...,\mathcal{D}_{S+T}\rbrace$. 
The $s$-th domain consisting of $N_s$ samples is represented as $\mathcal{D}_s=\lbrace(x^s_i,y^s_i)\rbrace_{i=1}^{N_s}$, where $x^s_i$ denotes the $i$-th sample and $y^s_i$ refers to its corresponding label.
Our goal is to leverage these source domains $\mathcal{S}$ to develop a model capable of seamlessly generalizing to any unseen target domain $\mathcal{T}$.

\textbf{First-order meta-learning} \cite{nichol2018first}
segments the optimization process into an inner loop and an outer loop.
Each iteration can be summarized as follows: a task comprises a batch of data sampled from a specific data distribution, and a step aggregates multiple tasks to form a larger batch.
In domain generalization, tasks are commonly partitioned by domains. 
MLDG \cite{li2018learning} evenly allocates tasks into two steps, with one step for the meta-train set and the other for the meta-test set. Fish \cite{shi2021gradient} selects one task for each step, which is demonstrated to achieve pairwise gradient matching between all domains.
During the inner loop, the model is sequentially updated with steps to reach parameters $\hat{\Theta}$.
Then in the outer loop, the original parameters $\Theta$ are updated towards $\hat{\Theta}$ in previous methods.

\textbf{Organization.} We present our arithmetic meta-learning in \cref{alg:arithmetic}. The parallel sections \cref{subsec:method-gradient} and \cref{subsec:method-arithmetic} compare between Arith and existing methods from two perspectives.
\cref{subsec:method-gradient} demonstrates we share the same properties of gradient matching as previous work. 
\cref{subsec:method-arithmetic} explains how our method achieves a more balanced positioning across source domains compared to prior approaches.

\begin{figure*}[t]
    \centering
    \includegraphics[width=0.9\linewidth]{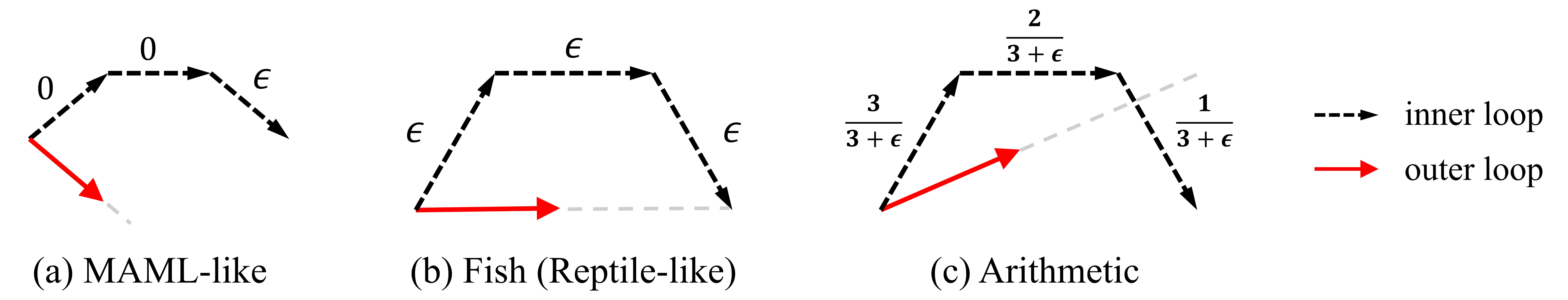}
    \caption{Comparison of different learning strategies. 
    Each step of the inner loop corresponds to a distinct domain, while in the outer loop, the gradient is computed as the weighted average of those from the inner loop, with the values above representing their respective weights.} 
    \vspace{-0.1in}
    \label{fig:compare}
\end{figure*}

\subsection{Gradient Matching}
\label{subsec:method-gradient}

We first introduce our proof process of gradient matching.
During an inner loop comprising $n$ steps, as the model's parameters transition from $\Theta$ to $\hat{\Theta}$, we depict the trajectory of parameter updates as $\lbrace \theta_1, \theta_2, ..., \theta_{n+1}\rbrace$, where $\theta_1$ and $\theta_{n+1}$ correspond to $\Theta$ and $\hat{\Theta}$, respectively.
At each step, the loss is denoted as $\lbrace \mathcal{L}_1, \mathcal{L}_2, ..., \mathcal{L}_n\rbrace$, and the corresponding gradients as $\lbrace g_1, g_2, ..., g_n\rbrace$, where $g_i = \alpha \nabla\mathcal{L}_i(\theta_i)$, with $\alpha$ indicating the learning rate of the inner loop.

First, let's represent the outer loop of existing methods $\Theta+\epsilon(\hat{\Theta}-\Theta)$ in gradient form: 
\begin{equation}
\label{eq:1}
\Theta \leftarrow \Theta-\epsilon \sum_{i=1}^n g_i,
\end{equation}
thus the optimization objective can be written as:
\begin{equation}
\label{eq:2}
\mathop{\rm argmin}_{\Theta}\, 
\sum_{i=1}^n\mathcal{L}_i(\theta_i).
\end{equation}
Given that each $g_i$ in \cref{eq:1} shares the common coefficient $\epsilon$, so the weights of loss $\mathcal{L}_i(\theta_i)$ in \cref{eq:2} are also the same.
$\mathcal{L}_i(\theta_i)$ can be approximated as the loss on the original parameters $\theta_1$ subtracting a regularization term. Please refer to our supplementary material for more details: 
\begin{equation}
\label{eq:3}
\mathcal{L}_i(\theta_i) = \mathcal{L}_i(\theta_1) - \alpha\sum_{j=1}^{i-1}\nabla\mathcal{L}_i(\theta_1) \cdot \nabla\mathcal{L}_j(\theta_1) + \mathcal{O}(\alpha^2).
\end{equation}
This equation aims to maximize dot product between gradient $i$ and those from the preceding $i-1$ steps, promoting smaller angles between gradients and thus ensuring consistency in the updating direction across domains.

The derivation above excludes the coefficient $\epsilon$ in \cref{eq:1} since the weights of losses in \cref{eq:2} can be arbitrary. 
This implies that as long as the update procedure is expressed as adjusting the original parameters $\Theta$ by the inner-loop gradients $\lbrace g_1, g_2, ..., g_n\rbrace$, gradient matching can be deduced, allowing for an infinite range of update directions to satisfy gradient matching.
The more general form is:
\begin{equation}
\label{eq:4}
\Theta \leftarrow \Theta- \sum_{i=1}^n \epsilon_i g_i,
\end{equation}
where $\epsilon_i$ are arbitrary coefficients.
For instance, if we consider averaging all intermediate models $\lbrace \theta_2, \theta_3, ..., \theta_{n+1}\rbrace$ during the inner loop, then the new outer loop is written as:
\begin{equation}
\label{eq:5}
\Theta \leftarrow \frac{1}{n+\epsilon}(\epsilon \theta_1 + \sum_{i=1}^{n}\theta_{i+1}).
\end{equation}
\cref{eq:5} represents the non-gradient form of our proposed arithmetic meta-learning. The optimization process can be obtained by substituting $\theta_{i+1} = \theta_1-\sum_{j=1}^{i}g_j$ into \cref{eq:5}, with gradient weights following an arithmetic progression that decreases from $\frac{n}{n+\epsilon}$ to $\frac{1}{n+\epsilon}$: 
\begin{equation}
\label{eq:6}
\Theta \leftarrow \Theta-\frac{1}{n+\epsilon} \sum_{i=1}^n (n+1-i) g_i.
\end{equation}
This form also adheres to \cref{eq:4}, illustrating that Arith can also achieve gradient matching as previous methods.

\begin{figure*}[t]
    \centering
    \includegraphics[width=0.9\linewidth]{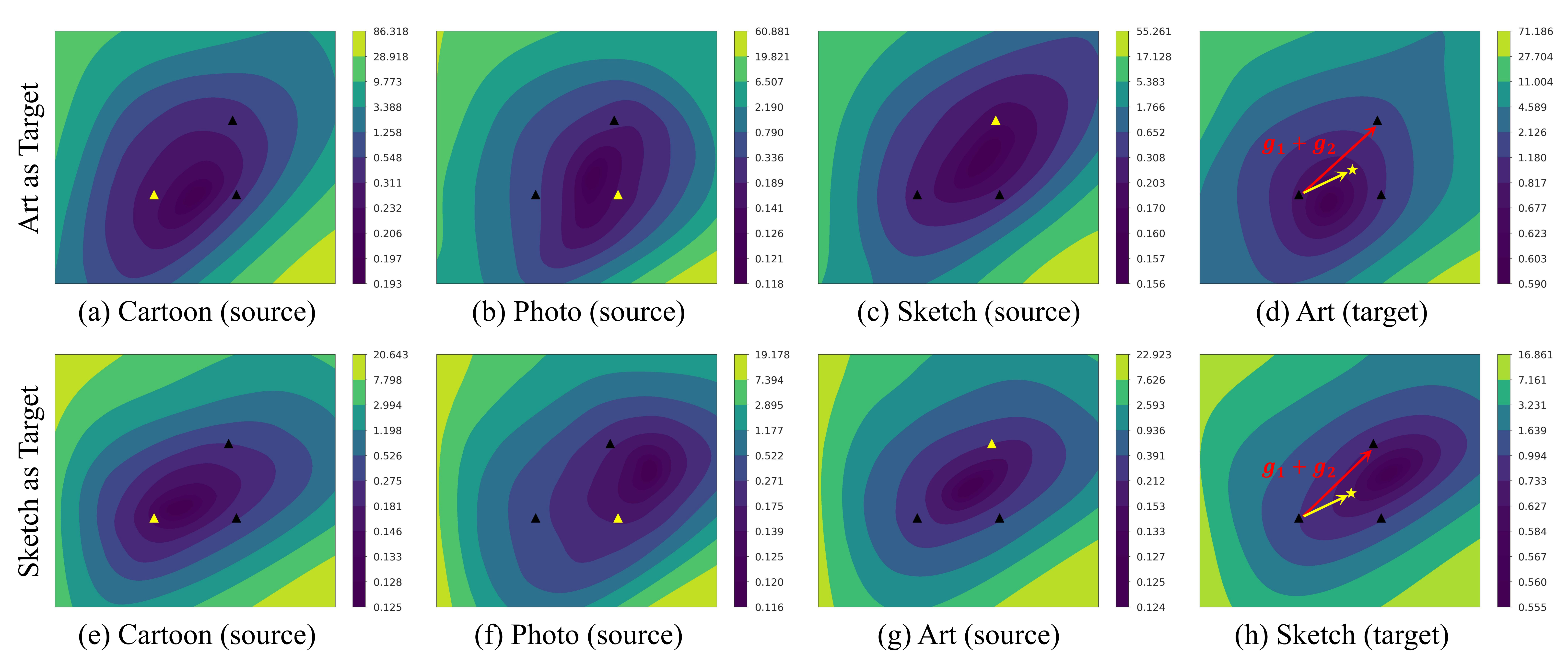}
    \caption{
    Loss surface plots of various domains on the PACS dataset, where \textbf{the deeper is better}. The yellow triangle in (a)(b)(c)(e)(f)(g) shows the estimated optimal parameters from the respective source domain, while the black triangle represents the estimated optimal parameters for the other source domains. The red arrow in (d)(h) is the updating direction of previous methods, while the \textbf{yellow arrow} towards the centroid marks the update direction of arithmetic meta-learning.
    }
    
    \vspace{-0.1in}
    \label{fig:surface}
\end{figure*}

\begin{figure}[t]
  \centering
  \includegraphics[width=1\linewidth]{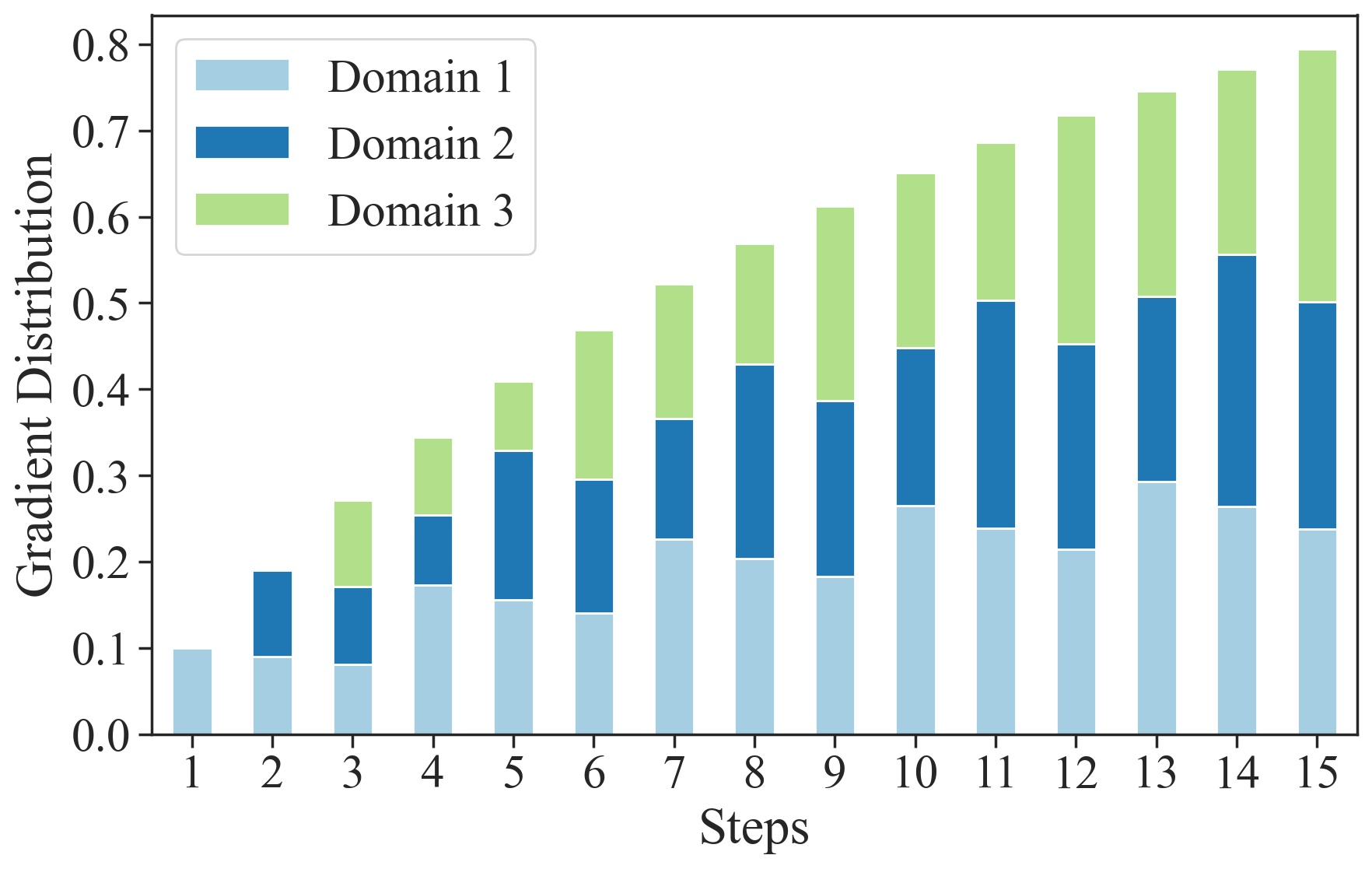}
   \caption{Adam optimizer's gradient distribution over the first fifteen steps of inner loop. Three domains are alternately optimized at each step. As momentum builds, the gradient contributions from each domain converge to similar proportions in the later stages. 
}
    \vspace{-0.1in}
    \label{fig:adam}
\end{figure}
\subsection{Balanced Positioning}
\label{subsec:method-arithmetic}

We adopt the analytical technique proposed in \cite{nichol2018first}.
Let $\mathcal{W}_i$ represent the optimal parameter set for domain $i$. Our goal is to determine $\Theta$ such that the distance $\mathcal{D}(\Theta, \mathcal{W}_i)$ is small and uniform among all source domains. We specify that each step is sampled from a single source domain $i$, then the step-wise optimization objective can be regarded as minimizing the squared distance: 
\begin{equation}
    \label{eq:7}
    \mathop{\rm argmin}_{\theta_i}\, \frac{1}{2}\mathcal{D}(\theta_i, \mathcal{W}_i)^2.
\end{equation}
Given a non-pathological set $\mathcal{Z} \subset \mathbb{R}^d$, for almost all points $\phi \in \mathbb{R}^d$
the gradient of the squared distance $\mathcal{D}(\phi, \mathcal{Z})^2$ equals $2(\phi-P_{\mathcal{Z}}(\phi))$, where $P_{\mathcal{Z}}(\phi)$ is the projection (\ie closest point) of $\phi$ onto $\mathcal{Z}$. Thus, the updating process at step $i$ can be represented as:
\begin{align}
    \theta_{i+1} &= \theta_i - \eta_i \nabla_{\theta_i}\frac{1}{2}\mathcal{D}(\theta_i, \mathcal{W}_i)^2 \\
    &= \theta_i - \eta_i (\theta_i-P_{\mathcal{W}_i}(\theta_i)) \\
    &= (1-\eta_i)\theta_i + \eta_i P_{\mathcal{W}_i}(\theta_i). \label{eq:10}
\end{align}
\cref{eq:10} illustrates that step $i$ can be interpreted as an interpolation between $\theta_i$ and its optimal projection onto the $i$-th source domain, where $\eta_i \in (0, 1)$ is not a hyperparameter, but a measure of their relative weights. \cref{eq:10} can also be iteratively expanded as:
\begin{equation}
    \label{eq:11}
    \theta_{i+1} = \prod_{j=1}^i(1-\eta_j) \cdot \theta_1 + \sum_{j=1}^{i} \prod_{k=j+1}^{i} (1-\eta_k) \cdot \eta_j \Phi_j.
\end{equation}
We denote $P_{\mathcal{W}_i}(\theta_i)$ as $\Phi_i$ to emphasize its optimality for domain $i$, regardless of which model $\theta_i$ it originates from. 
\cref{eq:11} reveals that $\theta_{i+1}$ is a weighted average of the initial model $\theta_1$ and the optimal parameters $\Phi$. 
In the second part of \cref{eq:11}, terms with smaller $j$ contain more $(1-\eta)$ that are smaller than $1$, indicating a progressive decay in the weights of earlier steps.
This confirms the tendency of models to prioritize the most recent data they encounter.

Previous methods typically interpolate between $\theta_1$ and $\theta_{n+1}$ during the outer loop. By applying \cref{eq:11}, we find that $\theta_1$ excludes $\Phi$, while $\theta_{n+1}$ has an unbalanced distribution of $\lbrace \Phi_1, \Phi_2, ..., \Phi_{n}\rbrace$. 
As a result, the interpolation between $\theta_1$ and $\theta_{n+1}$ also fails to correct this biased distribution.
In the early stages of training, this strategy may be acceptable if the model is far from $\Phi$, resulting in small values for $\eta$ so that $(1-\eta)$ is close to 1, which ensures relatively balanced weights for each $\Phi$.
However, as the model approaches convergence, oscillations around $\Phi$ cause $(1-\eta)$ to decrease, making it challenging to maintain equal weights for each domain-optimal parameters.

We propose to address this imbalance by utilizing the intermediate models $\lbrace \theta_2, \theta_3, ..., \theta_{n+1}\rbrace$ in the inner loop.
Each $\theta_{i+1}$ is expected to specifically tailored for its corresponding source domain $i$.
While accurately computing $\Phi$ is impractical, we can estimate it by performing multiple gradient updates on the respective source domain to minimize all $(1-\eta)$ towards $0$.
Ideally, only the last term in \cref{eq:11}, which exclusively involves $\eta$ without $(1-\eta)$ should be retained:
\begin{equation}
    \label{eq:12}
    \theta_{i+1} \approx \Phi_i.
\end{equation}
By substituting \cref{eq:12} into \cref{eq:5}, we derive:
\begin{equation}
\label{eq:13}
\Theta \leftarrow \frac{1}{n+\epsilon} (\epsilon\Theta + \sum_{i=1}^{n}\Phi_i). 
\end{equation}
This term represents the initial model $\Theta$ adjusted by the average of domain-optimal parameters, with each $\Phi$ equally weighted. Consequently, this strategy effectively achieves a good balance across source domains.

\begin{table*}[t]
\caption{Accuracy (\%) on the DomainBed benchmark across \textbf{five} datasets. The best and second-best results per dataset are \textbf{bolded} and \underline{underlined} respectively. Detailed results for each domain are available in the supplementary material.}
\vspace{-0.05in}
\centering
\begin{tabularx}{0.8\textwidth}{l>{\centering\arraybackslash}X>{\centering\arraybackslash}X>{\centering\arraybackslash}X>{\centering\arraybackslash}X>{\centering\arraybackslash}X>{\centering\arraybackslash}X}
\toprule
\textbf{Method} & {\textbf{PACS}} & {\textbf{VLCS}} & {\textbf{OfficeH}} & {\textbf{TerraInc}} & \textbf{DomainNet} & {\textbf{Avg}} \\
\midrule
ERM \cite{naumovich1998statistical} & 85.5 & 77.5 & 66.5 & 46.1 & 40.9 & 63.3 \\
IRM \cite{arjovsky2020invariant} & 83.5 & 78.6 & 64.3 & 47.6 & 33.9 & 61.6 \\
CausIRL \cite{chevalley2022invariant} & 83.6 & 76.5 & 68.1 & 47.4 & \underline{41.8} & 63.5 \\
VREx \cite{krueger2021out} & 84.9 & 78.3 & 66.4 & 46.4 & 33.6 & 61.9 \\
CORAL \cite{sun2016deep} & \underline{86.2} & \underline{78.8} & \underline{68.7} & 47.7 & 41.5 & \underline{64.5} \\
RSC \cite{huang2020self} & 85.2 & 77.1 & 65.5 & 46.6 & 38.9 & 62.7 \\
ARM \cite{zhang2020adaptive} & 85.1 & 77.6 & 64.8 & 45.5 & 35.5 & 61.7 \\
AND-mask \cite{parascandolo2020learning} & 84.4 & 78.1 & 65.6 & 44.6 & 37.2 & 62.0 \\
SAND-mask \cite{shahtalebi2021sand} & 84.6 & 77.4 & 65.8 & 42.9& 32.1 & 60.6 \\
MLDG \cite{li2018learning} & 84.9 & 77.2 & 66.8 & \underline{47.8} & 41.2 & 63.6 \\
Fish \cite{shi2021gradient} & 85.5 & 77.8 & 68.6 & 45.1 & \textbf{42.7} & 63.9 \\
Fishr \cite{rame2022fishr} & 85.5 & 77.8 & 67.8 & 47.4 & 41.7 & 64.0 \\
HGP \cite{hemati2023understanding} & 84.7 & 77.6 & 68.2 & 43.6 & 41.1 & 63.0 \\ 
Hutchinson \cite{hemati2023understanding} & 83.9 & 76.8 & 68.2 & 46.6 & 41.6 & 63.4 \\

\midrule

Arith & \textbf{86.5} \tiny{$\pm$ 0.3} & \textbf{79.4} \tiny{$\pm$ 0.3} & \textbf{69.4} \tiny{$\pm$ 0.1} & \textbf{48.1} \tiny{$\pm$ 1.2}  & 41.5 \tiny{$\pm$ 0.1} & \textbf{65.0} \\

\bottomrule

\end{tabularx}

\vspace{-0.1in}
\label{tab:domainbed}
\end{table*}
\subsection{Optimizer Selection}
\label{subsec:method-optimizer}

Previous research commonly (i) applies the Adam optimizer in the inner loop and (ii) performs direct interpolation in the outer loop. However, this method conflicts with the principle of domain-wise gradient matching, as the Adam optimizer contains momentum:
\begin{equation}
    \label{eq:14}
    \bar{g_{i}} = \beta \bar{g_{i-1}} + (1-\beta){g}_{i}, 
\end{equation}
where $\beta$ controls the weighting between momentum $\bar{g_{i-1}}$ and the current gradient $g_i$, with a default value of 0.9. Consequently, each $g_i$ in \cref{eq:1} is actually $\bar{g_i}$ in \cref{eq:14}, which becomes a blend of gradients from all domains. 
In \cref{fig:adam}, we visualize the gradients of Adam during alternating optimization across three domains for the first fifteen steps, illustrating how this gradient matching resembles uniform sampling without domain-specific gradients.

To prevent the failure of domain-wise gradient matching, we propose (i) using the Adam optimizer only in the outer loop and (ii) employing stochastic gradient descent without momentum in the inner loop to ensure each step accurately reflects the true gradient of its corresponding domain.

\subsection{Relationships with Weight Averaging}
\label{subsec:method-relationship}

Previous meta-learning methods for domain generalization closely resembles weight averaging \cite{izmailov2018averaging, cha2021swad}, as the interpolation between two models can be treated as the weighted average of them. 
For arithmetic meta-learning, we can also equivalently substitute it with the average of all intermediate models during the inner loop.
However, arithmetic meta-learning differs from averaging-based methods in several key aspects:
(i) Varied objectives: Model averaging aims to find flat minima across domain-agnostic models, ensuring robustness against shifts in the loss landscape between training and test sets. In contrast, arithmetic meta-learning prevents models from becoming overly biased towards specific domains by averaging between domain-specific models.
(ii) Different scope: Model averaging operates on a global scale, encompassing most of the models throughout the entire training process, while arithmetic meta-learning is more localized, averaging only a small subset of models during each iteration.
(iii) Unique implementation: By transforming weight averaging into gradient form, arithmetic meta-learning enables the optimization of outer loop to easily adapt to optimizers such as Adam.

Given that our strategy is based on the first-order meta-learning framework, it faces common challenges associated with such methods, including difficulties in converging to flat minima, and the balance between source domains may not translate to optimal parameters of the unseen target domain.
However, the orthogonal nature of arithmetic meta-learning and model averaging at local and global scales enables effective synergy between these approaches.. Experimental results in \cref{sec:experiment} further demonstrate that the combination of them leads to enhanced performance.

\section{Experiment}
\label{sec:experiment}

\subsection{Datasets}
\label{subsec:experiment:datasets}

We experiment on ten standard DG datasets, five of which are from the DomainBed \cite{gulrajani2020search} benchmark:
(i) \textbf{PACS} \cite{li2017deeper} contains 4 domains (\emph{photo}, \emph{art-painting}, \emph{cartoon}, \emph{sketch}) with 7 classes and 9,991 images. 
(ii) \textbf{Office-Home} \cite{venkateswara2017deep} comprises 4 domains (\emph{art}, \emph{clipart}, \emph{product}, \emph{real-world}) with 65 classes and 15,588 images.
(iii) \textbf{VLCS} \cite{fang2013unbiased} consists of 4 domains (\emph{pascal}, \emph{labelme}, \emph{caltech}, \emph{sun}) with 5 classes and 10,729 images.
(iv) \textbf{TerraIncognita} \cite{beery2018recognition} is composed of 4 domains (\emph{location38}, \emph{location43}, \emph{location46}, \emph{location100}) with 100 classes and 24,788 images.
(v) \textbf{DomainNet} \cite{peng2019moment} includes 6 domains (\emph{clipart}, \emph{infograph}, \emph{painting}, \emph{quickdraw}, \emph{real}, \emph{sketch}) with 345 classes and 586,575 images.

The other five datasets AMAZON, CAMELYON17 \cite{bandi2018detection}, CIVILCOMMENTS \cite{borkan2019nuanced}, IWILDCAM \cite{beery2020iwildcam}, and FMOW \cite{christie2018functional} are conducted on the WILDS \cite{koh2021wilds} benchmark which contains multiple modalities.
Due to space limitations, we present results only on CAMELYON17 in the main text and please refer to the supplementary material for others.

\begin{table}[t]
\caption{Ablation studies (\%) on \textbf{PACS} dataset. The best results are \textbf{bolded}. 
The $m$ denotes whether to enable momentum during the inner loop, while $ds$ signifies whether the data is partitioned by domains for each step or uniformly sampled.
}
\vspace{-0.05in}
\centering
\resizebox{1\linewidth}{!}{
\begin{tabular}{cccccccc}
\toprule
\textbf{Method} & \textbf{$m$} & $ds$ & {\textbf{A}} & {\textbf{C}} & {\textbf{P}} & {\textbf{S}} & {\textbf{Avg}} \\
\midrule
\multirow{3}{*}{MAML}  & - & - & 84.6 & 80.8 & 96.7 & 79.3 & 85.3 \\

& \checkmark & \checkmark & 84.8  & 81.9 & 96.0  & 77.5  & 85.0  \\

& - & \checkmark & 85.5 & 78.5  & \textbf{97.2}  & 76.4  & 84.4   \\ 

\midrule

\multirow{3}{*}{Fish} & - & 
- & 85.0  & 81.1  & 95.2 & 80.0  & 85.3 \\

& \checkmark & \checkmark &  85.6  & 82.0  & 95.7  & 78.5  & 85.4  \\ 

& - & \checkmark & 85.7  & 81.1  & 96.7  & 
81.0  & 86.1  \\

\midrule

\multirow{3}{*}{Arith} &
- & - & 85.6 & 80.7 & 96.3  & 80.9  & 85.9 \\

& \checkmark & \checkmark & 85.1  & \textbf{82.2}  & 96.6  & 78.9  & 85.7  \\

& - & \checkmark & \textbf{85.9} & 81.3  & 97.1  & \textbf{81.8}  & \textbf{86.5}  \\

\bottomrule
\end{tabular}
}
\vspace{-0.1in}
\label{tab:ablation}
\end{table}

\subsection{Implementation Details}
\label{subsec:experiment:implementation}

We follow the protocol proposed in DomainBed \cite{gulrajani2020search} and use ResNet50 \cite{he2016deep} pretrained on ImageNet \cite{deng2009imagenet} as our backbone network.
We apply stochastic gradient descent in the inner loop and the Adam optimizer in the outer loop, with each of the domains restricted to a single step over 5000 iterations.
The domains are optimized in random order, with a batch size of $32$ and a learning rate of $5e-5$.  
For datasets other than DomainNet, which consists of three domains, the weight of gradients are assigned as $\lbrace \sfrac{1}{2}, \sfrac{1}{3}, \sfrac{1}{6} \rbrace$.
For DomainNet, which comprises five domains, the corresponding weights are set as $\lbrace \sfrac{1}{3}, \sfrac{4}{15}, \sfrac{1}{5}, \sfrac{2}{15}, \sfrac{1}{15} \rbrace$.
We select one target domain for test and use the remaining domains for training and validation.
We reserve $20\%$ of the samples for validation from each source domain and choose the model with maximized accuracy on the overall validation set, which is the same as the \emph{training-domain validation set} in \cite{gulrajani2020search}.
Each experiment is conducted on a single Nvidia RTX 2080Ti GPU with Pytorch 1.10.1.

\begin{table}[t]
\caption{Ablation studies (\%) on \textbf{OfficeHome} and \textbf{VLCS} dataset. The best results are \textbf{bolded}. 
The scaled means increasing the learning rate to 1.5 times its original value, while adam refers to utilizing the Adam optimizer during the outer loop.
}
\vspace{-0.05in}
\centering
\resizebox{1\linewidth}{!}{
\begin{tabular}{cccccccc}
\toprule
\textbf{Method} & scaled & adam & {\textbf{D1}} & {\textbf{D2}} & {\textbf{D3}} & {\textbf{D4}} & {\textbf{Avg}} \\
\midrule

\multicolumn{8}{c}{OfficeHome} \\

\midrule

\multirow{4}{*}{Fish} & \checkmark & 
- & 61.5  & 55.3  & 74.6 & 77.2  & 67.2 \\

& \checkmark & \checkmark &  \textbf{65.0}  & 54.8  & 77.0  & 78.8  & 68.9  \\ 

& - & - & 60.7  & 53.6  & 75.7  & 
77.3  & 66.8  \\

& - & \checkmark & 64.3  & 55.3  & 77.2  & 
79.0  & 69.0  \\

\midrule

\multirow{4}{*}{Arith} &
\checkmark & - & 62.5 & 54.7 & 76.4  & 77.2  & 67.7 \\

& \checkmark & \checkmark & 64.8  & 55.8  & 76.1  & 
\textbf{79.5}  & 69.1  \\

& - & - & 61.6  & 55.2  & 75.6  & 77.1  & 67.4  \\

& - & \checkmark & 64.7 & \textbf{56.3}  & \textbf{77.5}  & 79.2  & \textbf{69.4}  \\

\midrule

\multicolumn{8}{c}{VLCS} \\

\midrule

\multirow{4}{*}{Fish} & \checkmark & 
- & 97.1  & 63.7  & 72.3 & 77.9  & 77.8 \\

& \checkmark & \checkmark &  98.6  & 64.3  & \textbf{76.7}  & 76.4  & 79.0 \\ 

& - & - & 97.5  & 63.8  & 73.6  & 
78.1  & 78.3  \\

& - & \checkmark & 98.7  & \textbf{65.0}  & 76.5  & 
76.0  & 79.2  \\

\midrule

\multirow{4}{*}{Arith} &
\checkmark & - & 97.3 & 64.8 & 73.2  & 77.8  & 78.3 \\

& \checkmark & \checkmark & 98.7  & 64.1  & 76.3  & 
\textbf{78.2}  & 79.3  \\

& - & - & 98.5  & 64.7  & 76.0  & 77.3  & 79.1  \\

& - & \checkmark & \textbf{98.7} & 64.6  & 76.3  & 77.8  & \textbf{79.4}  \\ 

\bottomrule
\end{tabular}
}
\vspace{-0.1in}
\label{tab:ablation-2}
\end{table}
\subsection{Main Results}
\label{subsec:experiment:results}

As illustrated in \cref{tab:domainbed}, we first compare arithmetic meta-learning with classic domain generalization methods on the DomainBed \cite{gulrajani2020search} benchmark, where the strategies below RSC \cite{huang2020self} are gradient-based.
For each dataset, we perform experiments three times and report the average results followed by the standard deviation.
Our method demonstrates superior performance on four datasets, surpassing the second-best method by $0.3\%$, $0.6\%$, $0.7\%$ and $0.3\%$ respectively.
We attribute our improvement to two key aspects. 
First, our approach can estimate the centroid among domain experts, which still offers opportunities for further enhancement in this table. Since our method requires precise estimation of optimal parameters for each source domain, updating parameters only once may introduce bias. To address this, we conduct multiple-step per domain experiments in the following ablation studies.
Second, we refine the usage of optimizers from previous methods. By removing momentum during the inner loop, we achieve more precise gradient matching across domains, while using the Adam optimizer in the outer loop helps ensure smoother convergence for the final model.

\subsection{Ablation \& Analysis}
\label{subsec:experiment:analysis}

\textbf{Loss surface visualization.}
We visualize the loss surface using the technique proposed in \cite{izmailov2018averaging}.
We select three intermediate models from the inner loop and compute their linearly interpolated losses, where each model is sequentially derived from 30 gradient updates on a single source domain. 
We calculate the averaged parameters of these models (\ie, yellow stars), which represent the optimization direction of arithmetic meta-learning in the outer loop.
As shown in \cref{fig:surface}, while the loss basins differ between domains, the variations are not substantial. 
The relative positions of each source domain's loss basins closely align with those of their corresponding intermediate models (\ie, yellow triangles), suggesting their capacity to approximate domain-optimal parameters, thus the averaged parameters can be regarded as a good balance across source domains.
However, although the final averaged model happens to match with the basin of the target domain in \cref{fig:surface}, it's essential to acknowledge that since the target domain is unseen and arbitrary, our ability is limited to preventing overfitting to source domains, but cannot guarantee optimal performance on the target domain in every scenario.

\begin{table}[t]
\caption{Ablation study (\%) with global averaging of \textbf{five} datasets on the DomainBed benchmark. The best results are \textbf{bolded}.
}
\vspace{-0.05in}
\centering
\begin{tabularx}{1\linewidth}{l>{\centering\arraybackslash}X>
{\centering\arraybackslash}X>{\centering\arraybackslash}X>{\centering\arraybackslash}X>{\centering\arraybackslash}X>{\centering\arraybackslash}X>{\centering\arraybackslash}X}
\toprule
\textbf{Method} & swad & {\textbf{P}} & {\textbf{V}} & {\textbf{O}} & {\textbf{T}} & \textbf{D} & {\textbf{Avg}} \\
\midrule

\multirow{2}{*}{ERM} & - &  84.7 & 78.0 & 67.8 & 46.5 & 40.8 & 63.6 \\
& \checkmark & 87.5 & 78.1 & \textbf{70.4} & 49.4 & 44.1 & 65.9 \\

\midrule

\multirow{2}{*}{MLDG} & - & 85.4 & 77.8 & 68.5 & 47.3 & 41.4 & 64.1 \\
& \checkmark & 88.3 & 78.9 & 69.8 & \textbf{50.5} & 44.6 & 66.4 \\

\midrule

\multirow{2}{*}{Fish} & - & 86.1 & 78.4 & 68.3 & 47.0 & 41.5 & 64.3  \\
& \checkmark & 88.5 & 79.4 & 70.1 & 49.5 & 44.9 & 66.5 \\ 

\midrule

\multirow{2}{*}{Arith} & - & 86.5 & 79.4 & 69.4  & 48.1   & 41.5 & 65.0 \\

& \checkmark & \textbf{88.7}  & \textbf{79.8}  & 70.2 & 49.9 & \textbf{44.9} & \textbf{66.7}  \\

\bottomrule

\end{tabularx}

\vspace{-0.1in}
\label{tab:swad}
\end{table}

\begin{figure}[t]
  \centering
  \includegraphics[width=0.85\linewidth]{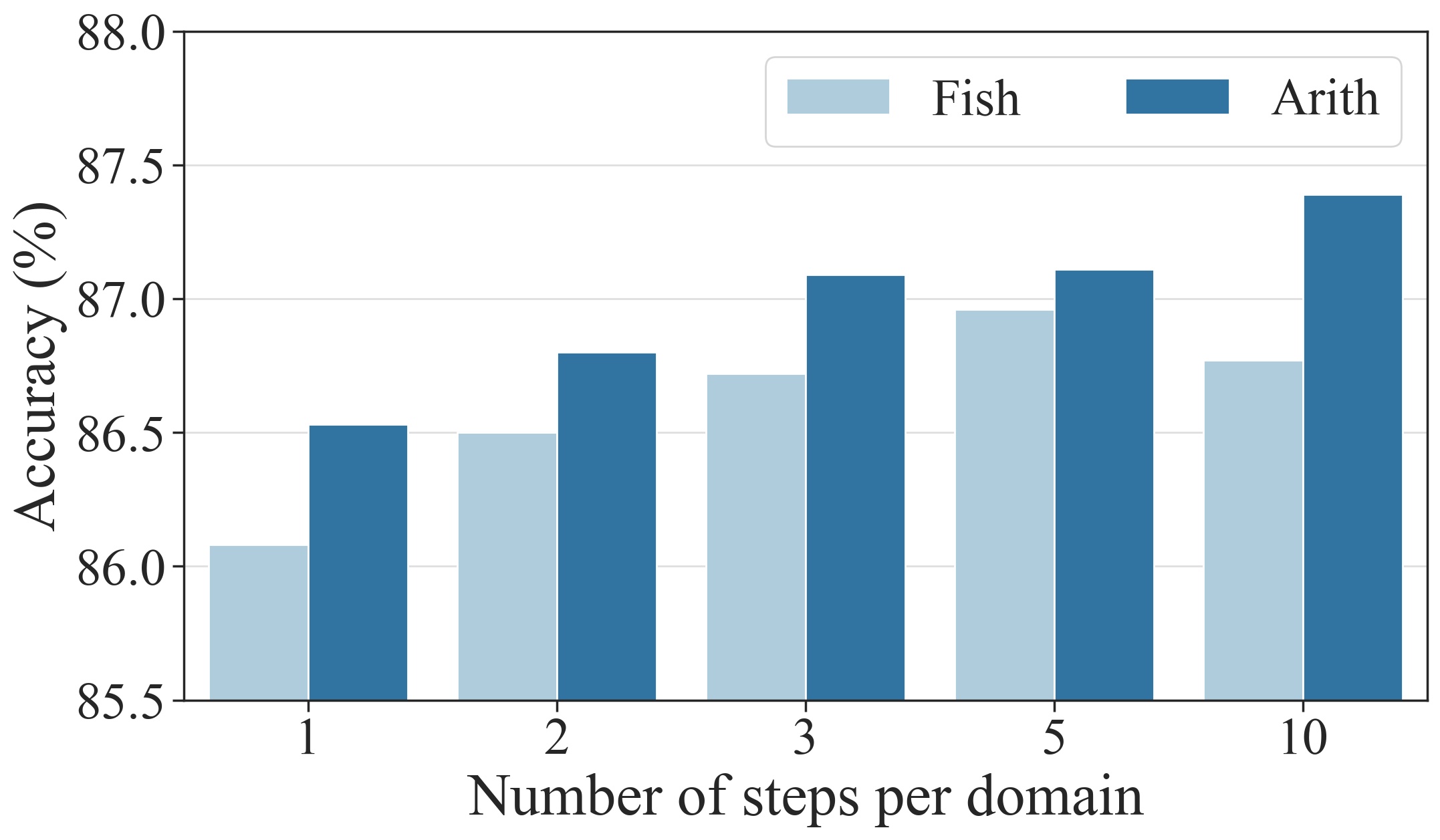}
   \caption{Accuracy (\%) on \textbf{PACS} dataset with the varying number of steps for each domain during the inner loop . 
}
    \vspace{-0.1in}
    \label{fig:step_variance}
\end{figure}

\textbf{Ablation study on steps per domain $k$.} 
We investigate how the performance correlates with the step count for each domain. 
As illustrated in \cref{fig:step_variance}, the optimization process of Fish \cite{shi2021gradient} signify linear interpolations between the initial and final parameters, while the arithmetic gradient weights simulates the updating direction towards the average of all intermediate models. 
As the number of steps increases, the accuracy of arithmetic meta-learning consistently improves. This indicates that performing multiple gradient updates results in a more precise estimation of optimal parameters for each domain, achieving a better balance through averaging. In contrast, the performance of Fish initially improves but then declines due to its biased estimation of the optimal balance. As the displacement of parameters increases, this inaccuracy may become more serious.

\textbf{Ablation study on gradient weights / optimizer selection.} 
The weights of gradients in the inner loop have two key attributes: ratio and magnitude. The ratio determines the final update direction and is fixed for each strategy. For instance, a $1:1:1$ ratio for Fish means updating in the direction of the last model, while a $3:2:1$ ratio for Arith means shifting towards the average of domain experts.
The magnitude affects the step size of the outer loop, similar to the learning rate.
To match with inner loop, we set the total sum of weights around 1, such as $\lbrace \sfrac{1}{3}, \sfrac{1}{3}, \sfrac{1}{3} \rbrace$ for Fish and $\lbrace \sfrac{1}{2}, \sfrac{1}{3}, \sfrac{1}{6} \rbrace$ for Arith. 
In DomainBed, Fish is implemented with weights of $\lbrace \sfrac{1}{3}, \sfrac{1}{3}, \sfrac{1}{3} \rbrace$, so we additionally set Arith's weights to $\lbrace \sfrac{3}{4}, \sfrac{1}{2}, \sfrac{1}{4} \rbrace$ when scaling is enabled, ensuring that the sum of the weights is equal.
As shown in \cref{tab:ablation-2}, Arith generally outperforms Fish in most scenarios and achieves higher average accuracy. 
Unlike the inner loop, the outer loop optimizer permits momentum.
When using the Adam optimizer, the impact of gradient weights on model performance is minimal due to Adam’s adaptive learning rate, which leads to better outcomes than models that do not utilize this condition.

\textbf{Ablation study on domain-specific steps / gradients.}
Gradient-based meta-learning methods for domain generalization typically select samples from each domain to perform domain-specific gradient matching. We also conduct experiments with uniform sampling and inner loop momentum. 
As shown in \cref{tab:ablation}, domain-specific sampling strategies without momentum generally outperform other implementations, while the performance of uniform sampling and momentum is comparable.
It is noted in \cref{subsec:method-optimizer} that uniform sampling and inner loop momentum results in mixed gradients, which fail to effectively match across source domains.
In arithmetic meta-learning, parameters obtained through domain-specific gradient matching are more likely to estimate globally optimal parameters rather than being tailored to individual ones, thus mitigating the risk of overfitting towards specific domains.

\textbf{Combining with global averaging.}
Arithmetic meta-learning simulates localized model averaging during each iteration, aiming to find a balanced position among source domains instead of pursuing a global flat minima. To further enhance this approach, we propose to integrate it with the global averaging method \cite{cha2021swad}, which facilitates the achievement of well-generalized flat minima. As shown in \cref{tab:swad}, our method yields additional improvements over this strong baseline, underscoring the complementary strengths of local and global averaging techniques. It is important to note that some of the results differ from those in \cref{tab:domainbed}, as they are based on our own implementation.

\section{Related Work}
\label{sec:related}
\begin{table*}[h]
\caption{Accuracy (\%) on \textbf{CAMELYON17} dataset. The best results are \textbf{bolded}.}
\vspace{-0.05in}
\centering
\begin{tabular}{c|ccccccccccc}
\toprule
\textbf{Method} & \textbf{20} & \textbf{21} & \textbf{22} & \textbf{23} & \textbf{24} & \textbf{25} & \textbf{26} & \textbf{27} & \textbf{28} & \textbf{29} & \textbf{Avg} \\
\midrule
ERM & 49.2 & 30.2 & 73.6 & 74.8 & 64.4 & 60.8 & 57.0 & 37.8 & 89.6 & 77.3 & 73.1 \\
Fish & 52.4 & \textbf{36.0} & 72.3 & \textbf{77.5} & 69.0 & 65.1 & 59.3 & \textbf{43.6} & 90.0 & 77.6 & 74.8 \\
Arith & \textbf{54.4} & 33.8 & \textbf{83.6} & 75.2 & \textbf{72.5} & \textbf{69.5} & \textbf{64.0} & 40.7 & \textbf{90.1} & \textbf{79.9} & \textbf{76.6} \\

\bottomrule
\end{tabular}
\vspace{-0.05in}
\label{tab:camelyon17}
\end{table*}
\subsection{Domain Generalization}
\label{subsec:related-dg}
One key aspect of domain generalization is to learn features invariant across source domains, assuming they will generalize to unseen target domains as well. Approaches such as domain adversarial learning \cite{dayal2024madg, sicilia2023domain} aim to train a feature extractor insensitive to domain-specific traits.
Invariant risk minimization \cite{arjovsky2020invariant, ahuja2021invariance} enforces consistency of the optimal classifier across all domains within the representation space.
Feature decoupling methods aim to \cite{chattopadhyay2020learning, lv2022causality} distill domain-invariant semantic information from original features.
To diversify training data and improve generalized feature representations, data augmentation methods manipulate statistical characteristics at either the image or feature level.
Commonly employed techniques include mixing \cite{zhou2020domain, wang2020heterogeneous} and Fourier transformations \cite{xu2021fourier, guo2023aloft}.
Rather than strictly enforcing an invariance factor, balance-based methods typically propose model-agnostic strategies to reduce overfitting to specific source domains.
Meta-learning \cite{dou2019domain,  zhang2022mvdg} aims to leverage prior knowledge to guide the learning of current tasks.  
Ensemble learning \cite{zhou2021domain, li2022domain} presumes the target domain distribution as a blend of source domain distributions, thereby incorporating statistics from domain expert models. Weight averaging \cite{cha2021swad, arpit2022ensemble} aggregates model weights across training episodes to achieve flatter minima that are less prone to overfitting.

\subsection{Meta-Learning}
\label{subsec:related-ml}

Meta-learning \cite{thrun2012learning, al2021data} is a well-established field aimed at enabling models to generalize across a diverse range of tasks.
Initially, meta-learning focused on discovering initial parameters that can adapt swiftly to new tasks within a few gradient steps.
For instance, model-agnostic meta-learning (MAML) \cite{finn2017model} and first-order meta-learning (Reptile) \cite{nichol2018first} split the optimization process into inner and outer loops. The inner loop handles task-specific adaptation, while the outer loop seeks a globally optimal initialization.
More recently, meta-learning has been applied to domain generalization (MLDG) \cite{li2018learning} to simulate domain shifts by synthesizing meta-train and meta-test domains. 
S-MLDG \cite{li2020sequential} enhances MLDG with sequential and lifelong learning. To address the computational cost of second-order derivatives, Fish \cite{shi2021gradient} use first-order algorithms for efficient and fine-grained task sampling.
Traditional meta-learning emphasizes intra-task gradient matching, whereas meta-learning for domain generalization focuses on inter-domain gradient matching.
Despite differing objectives, both approaches have demonstrated strong performance in their respective fields. We believe meta-learning for domain generalization can inform traditional meta-learning, given the conceptual parallel between domains and tasks. Supporting research \cite{lee2021sequential} also indicates that inter-task gradient matching contributes to effective initialization for general  tasks.
\subsection{Weight Averaging}
\label{subsec:related-wa}

Weight averaging \cite{izmailov2018averaging, wortsman2022model} is a variant of ensemble learning \cite{dong2020survey, yang2023survey}, in which the outputs of individual models are combined to enhance overall performance. However, traditional ensemble methods incur high computational and storage costs due to the need for training and inference with multiple models. Weight averaging addresses this issue by merging these models into a single entity. By averaging parameters from similar training trajectories, $\lbrace \theta_1, \theta_2, ..., \theta_{n}\rbrace$, previous work has demonstrated that it effectively approximates the ensemble output \cite{izmailov2018averaging, arpit2022ensemble}:
\begin{equation}
    f(\frac{1}{n}\sum_{i=1}^n \theta_i) \approx \frac{1}{n}\sum_{i=1}^n f(\theta_i).
\end{equation}
SWA \cite{izmailov2018averaging} uses a specialized learning rate schedule and periodically aggregates model weights. SWAD \cite{cha2021swad} imposes constraints on the sampling range and increases the sampling frequency to address overfitting in domain generalization tasks. Some methods focus on model selection \cite{wortsman2022model, rame2022diverse} by using a greedy algorithm to iteratively select the most effective model during the averaging process. Methods incorporating multiple updating trajectories are also employed in weight averaging \cite{jolicoeur2023population, arpit2022ensemble}, such as aligning the directions or leveraging the combined model outputs from different trajectories to achieve improved performance.
\section{Conclusion}
\label{sec:conclusion}

In this paper, we propose a simple yet effective enhancement to first-order meta-learning for domain generalization. While previous methods rely on constant gradient weights in the inner loop to ensure gradient matching across source domains, they overlook the crucial positioning of the model relative to the optimal parameters of each domain.
We introduce an arithmetic gradient-based meta-learning strategy that approximates the direction toward the average of all models within the inner loop. While preserving the principle of gradient matching, our approach guides the model towards the centroid of the each domain-optimal parameters, achieving a more precise balance.
Experimental results demonstrate the superior performance of our method.



\end{document}